# Probabilistic Logic Programming under Inheritance with Overriding

Thomas Lukasiewicz
Institut und Ludwig Wittgenstein Labor für Informationssysteme, TU Wien
Favoritenstraße 9-11, A-1040 Vienna, Austria
lukasiewicz@kr.tuwien.ac.at

## Abstract

We present probabilistic logic programming under inheritance with overriding. This approach is based on new notions of entailment for reasoning with conditional constraints, which are obtained from the classical notion of logical entailment by adding inheritance with overriding. This is done by using recent approaches to probabilistic default reasoning with conditional constraints. We analyze the semantic properties of the new entailment relations. We also present algorithms for probabilistic logic programming under inheritance with overriding, and we analyze its complexity in the propositional case.

## 1 INTRODUCTION

A number of recent research efforts are directed towards integrating logic-oriented and probability-based representation and reasoning formalisms. In particular, there are approaches to *probabilistic logic programming* that combine logic programming techniques with probabilities over possible worlds [25, 26, 4, 5, 19]. They are based on the model-theoretic notion of logical entailment, which is well-known from probabilistic propositional logics [28, 7, 6].

The notion of logical entailment, however, has often been criticized in the literature for its inferential weakness. For this reason, many recent approaches towards integrating logic and probabilities combine logic-based formalisms with Bayesian networks [33, 32, 12, 27, 14].

Another way to overcome the inferential weakness of logical entailment is to use the principle of maximum entropy [24] or the principle of sequential maximum entropy [20], where the latter is closely related to Bayesian networks. The maximum entropy approach, however, has the drawback that it does not properly model imprecision in our knowledge base. That is, maximum entropy always produces a single joint distribution, for example, also in the extreme case when our knowledge base is empty.

In this paper, we investigate a very promising new approach of strengthening the notion of logical entailment in probabilistic logic, which does not have the above-mentioned drawback of the maximum entropy approach. This approach also has advantages over the above combinations of logic-based formalisms with Bayesian networks, as it does not assume acyclic Bayesian network structures with complete and precise conditional probabilities.

This new approach is inspired by *reference-class reasoning*, which goes back to Reichenbach [34] and was further refined especially by Kyburg [16, 17] and Pollock [31].

Reichenbach [34] describes reference-class reasoning as follows: "If we are asked to find the probability holding for an individual future event, we must first incorporate the case in a suitable reference class. An individual thing or event may be incorporated in many reference classes .... We then proceed by considering the narrowest reference class for which suitable statistics can be compiled."

That is, Reichenbach suggests to equate our knowledge about a particular individual with the statistics from a reference class, which is informally defined as a set of individuals to which our particular individual belongs and about which we have "suitable statistics". Moreover, if there are several reference classes with conflicting statistics, then we should prefer the smallest one and its associated statistics.

Interestingly, Reichenbach's guidelines may be interpreted as *inheritance with overriding* as it is known from object-oriented programming languages. The aspect of *inheritance* is expressed by the fact that any class containing the particular individual can be considered as reference class, while the aspect of *overriding* is expressed by the fact that smaller reference classes are preferred to larger ones.

It turns out that the classical notion of logical entailment in probabilistic logic does not follow the principle of inheritance with overriding. It is thus a natural idea to strengthen logical entailment by adding inheritance with overriding. In this paper, we realize this idea by using recent ap-proaches to *probabilistic default reasoning* from [21, 22].

The main contributions of this paper are as follows:

- We present probabilistic logic programming under inheritance with overriding, which is based on recent approaches to probabilistic default reasoning from [21, 22].

- We describe some general nonmontonic properties of entailment under inheritance with overriding.

- We present algorithms for probabilistic logic programming under inheritance with overriding.

- We analyze the propositional complexity of probabilistic logic programming under inheritance with overriding.

Note that all proofs are given in the extended paper [23].

## 2 PRELIMINARIES

### 2.1 PROBABILISTIC BACKGROUND

We briefly recall how first-order logics of probability are given a semantics in which probabilities are defined over a set of possible worlds (cf. especially [3, 8, 35, 13]). We restrict our considerations to a language of first-order Boolean combinations of conditional constraints that are implicitly universally quantified and that are interpreted by probabilities over a set of Herbrand interpretations.

Let $\Phi$ be a first-order vocabulary that contains a finite set of predicate symbols and a finite set of constant symbols. Let $\mathcal{X}$ be a set of *object* and *bound variables*. Object variables represent elements of a certain domain, while bound variables describe real numbers in the unit interval $[0, 1]$.

An *object term* is a constant symbol from $\Phi$ or an object variable from $\mathcal{X}$. We define *classical formulas* by induction as follows. The propositional constants *false* and *true*, denoted $\bot$ and $\top$, respectively, are classical formulas. If $p$ is a predicate symbol of arity $k \geq 0$ from $\Phi$ and $t_1, \ldots, t_k$ are object terms, then $p(t_1, \ldots, t_k)$ is a classical formula (called *atom*). If $\phi$ and $\psi$ are classical formulas, then also $\neg \phi$ and $(\phi \wedge \psi)$. A *conditional constraint* is an expression of the form $(\psi|\phi)[l, u]$ with real numbers $l, u \in [0, 1]$ and classical formulas $\phi$ and $\psi$. We define *probabilistic formulas* inductively as follows. Every conditional constraint is a probabilistic formula. If $F$ and $G$ are probabilistic formulas, then also $\neg F$ and $(F \wedge G)$. We use $(F \vee G)$, $(F \Leftarrow G)$, and $(F \Leftrightarrow G)$ to abbreviate $\neg(\neg F \wedge \neg G)$, $\neg(\neg F \wedge G)$, and $(\neg(\neg F \wedge G) \wedge \neg(F \wedge \neg G))$, respectively, and adopt the usual conventions to eliminate parentheses. Object terms and formulas are *ground* iff they do not contain any variables. The notions of substitutions and of ground instances of probabilistic formulas are canonically defined.

We divide conditional constraints into *classical conditional constraints*, which have the form $(\psi|\phi)[1, 1]$ or $(\psi|\phi)[0, 0]$, and *purely probabilistic conditional constraints*, which are of the form $(\psi|\phi)[l, u]$ with $l < 1$ and $u > 0$.

We use $HB_\Phi$ (resp., $HU_\Phi$) to denote the Herbrand base (resp., Herbrand universe) over $\Phi$. In the sequel, we assume that $HB_\Phi$ is nonempty. A *possible world* $I$ is a subset of $HB_\Phi$. We use $\mathcal{I}_\Phi$ to denote the set of all possible worlds over $\Phi$. A *variable assignment* $\sigma$ maps each object variable to an element of $HU_\Phi$, and each bound variable to a real number from $[0, 1]$. It is extended to object terms by $\sigma(c) = c$ for all constant symbols $c$ from $\Phi$. The *truth* of classical formulas $\phi$ in $I$ under $\sigma$, denoted $I \models_\sigma \phi$, is inductively defined as follows (we write $I \models \phi$ when $\phi$ is ground):

- $I \models_\sigma p(t_1, \ldots, t_k)$ iff $p(\sigma(t_1), \ldots, \sigma(t_k)) \in I$.
- $I \models_\sigma \neg \phi$ iff not $I \models_\sigma \phi$.
- $I \models_\sigma (\phi \wedge \psi)$ iff $I \models_\sigma \phi$ and $I \models_\sigma \psi$.

A *probabilistic interpretation* $Pr$ is a probability function on $\mathcal{I}_\Phi$ (that is, since $\mathcal{I}_\Phi$ is finite, simply a mapping from $\mathcal{I}_\Phi$ to the unit interval $[0, 1]$ such that all $Pr(I)$ with $I \in \mathcal{I}_\Phi$ sum up to 1). The *probability* of a classical formula $\phi$ in the probabilistic interpretation $Pr$ under a variable assignment $\sigma$, denoted $Pr_\sigma(\phi)$ (or simply $Pr(\phi)$ when $\phi$ is ground), is defined as the sum of all $Pr(I)$ such that $I \in \mathcal{I}_\Phi$ and $I \models_\sigma \phi$. For classical formulas $\phi$ and $\psi$ with $Pr_\sigma(\phi) > 0$, we use $Pr_\sigma(\psi|\phi)$ to abbreviate $Pr_\sigma(\psi \wedge \phi) / Pr_\sigma(\phi)$. The *truth* of a probabilistic formula $F$ in a probabilistic interpretation $Pr$ under a variable assignment $\sigma$, denoted $Pr \models_\sigma F$, is inductively defined by:

- $Pr \models_\sigma (\psi|\phi)[l, u]$ iff $Pr_\sigma(\phi) = 0$ or $Pr_\sigma(\psi|\phi) \in [l, u]$.
- $Pr \models_\sigma \neg F$ iff not $Pr \models_\sigma F$.
- $Pr \models_\sigma (F \wedge G)$ iff $Pr \models_\sigma F$ and $Pr \models_\sigma G$.

A probabilistic formula $F$ *is true* in $Pr$, or $Pr$ is a *model* of $F$, denoted $Pr \models F$, iff $Pr \models_\sigma F$ for all variable assignments $\sigma$. We say $Pr$ is a *model* of a set of probabilistic formulas $\mathcal{F}$, denoted $Pr \models \mathcal{F}$, iff $Pr$ is a model of all $F \in \mathcal{F}$. We say $\mathcal{F}$ is *satisfiable* iff a model of $\mathcal{F}$ exists.

We next define the notion of *logical entailment* as follows. A probabilistic formula $F$ is a *logical consequence* of a set of probabilistic formulas $\mathcal{F}$, denoted $\mathcal{F} \models F$, iff each model of $\mathcal{F}$ is also a model of $F$. A conditional constraint $(\psi|\phi)[l, u]$ is a *tight logical consequence* of $\mathcal{F}$, denoted $\mathcal{F} \models_{tight} (\psi|\phi)[l, u]$, iff $l$ (resp., $u$) is the infimum (resp., supremum) of $Pr_\sigma(\psi|\phi)$ subject to all models $Pr$ of $\mathcal{F}$ and all variable assignments $\sigma$ with $Pr_\sigma(\phi) > 0$. Note that we assume $l = 1$ and $u = 0$, when $\mathcal{F} \models (\phi|\top)[0, 0]$ (that is, $Pr_\sigma(\phi) = 0$ for all models $Pr$ of $\mathcal{F}$ and all $\sigma$).

### 2.2 PROBABILISTIC LOGIC PROGRAMS

A *(general) probabilistic logic program* $P$ is a finite set of conditional constraints $(\psi|\phi)[l, u]$ with $l \leq u$. The *grounding* of $P$, denoted $ground(P)$, is the set of all ground instances of members of $P$. A *probabilistic query* is an expression of the form $\exists(\beta|\alpha)[s, t]$, where $\alpha$ and $\beta$ are two classical formulas, and $s$ and $t$ are either two real numbers from $[0, 1]$ or two distinct bound variables from $\mathcal{X}$. It is *object-ground* iff $\alpha$ and $\beta$ are ground and $s, t \in \mathcal{X}$.

We say $(\psi|\phi)[l, u]$ is *conjunctive* (resp., *1-conjunctive*) iff $\psi$ is a conjunction of atoms (resp., $\psi$ is an atom) and $\phi$ is either $\top$ or a conjunction of atoms. A probabilistic logic program $P$ is *conjunctive* (resp., *1-conjunctive*) iff all $C \in P$ are conjunctive (resp., 1-conjunctive). A *probabilistic query* $\exists(\beta|\alpha)[s, t]$ is *conjunctive* (resp., *1-conjunctive*) iff $(\beta|\alpha)[0, 1]$ is conjunctive (resp., 1-conjunctive).

The meaning of probabilistic queries to probabilistic logic programs is defined by entailment semantics for probabilistic logic programs. Every semantics $s$ is associated with an *s-consequence* relation $\|\sim^s$ and a *tight s-consequence* relation $\|\sim^s_{tight}$, which are subsets of $2^{\mathcal{L}_\Phi} \times \mathcal{L}_\Phi$, where $\mathcal{L}_\Phi$ denotes the set of all conditional constraints over $\Phi$.

Two entailment semantics based on logical entailment, called *0-* and *1-entailment*, are defined as follows. The *0-consequence* (resp., *tight 0-consequence*) relation is given by $\models$ (resp., $\models_{tight}$). Note that reasoning in probabilistic logics is in general done with 0-entailment. A conditional constraint $(\psi|\phi)[l, u]$ is a *1-consequence* of a set of conditional constraints $C$ iff $Pr_\sigma(\psi) \in [l, u]$ for all models $Pr$ of $C$ and all variable assignments $\sigma$ such that $Pr_\sigma(\phi) = 1$. We say $(\psi|\phi)[l, u]$ is a *tight 1-consequence* of $C$ iff $l$ (resp., $u$) is the infimum (resp., supremum) of $Pr_\sigma(\psi)$ subject to all models $Pr$ of $C$ and all $\sigma$ such that $Pr_\sigma(\phi) = 1$.

The main difference between 0- and 1-entailment is that 0-entailment is based on *conditioning*, while 1-entailment realizes some *constraining*. For example, a ground conditional constraint $(\psi|\phi)[l, u]$ is a 0-consequence of $C$ iff $Pr[\phi](\psi) \in [l, u]$ for every model $Pr$ of $C$ with $Pr(\phi) > 0$, where $Pr[\phi]$ denotes the conditioning of $Pr$ on $\phi$. Whereas $(\psi|\phi)[l, u]$ is a 1-consequence of $C$ iff $Pr(\psi) \in [l, u]$ for every model $Pr$ of $C$ with $Pr(\phi) = 1$. Note that under 0- and 1-entailment, probabilistic logic programs $P$ are equivalent to their groundings $ground(P)$.

Given a probabilistic query $\exists(\beta|\alpha)[l, u]$ with $l, u \in [0, 1]$ to a probabilistic logic program $P$, its *correct answer substitutions under* a semantics $s$ are substitutions $\theta$ such that $P \|\sim^s (\beta\theta|\alpha\theta)[l, u]$ and that $\theta$ acts only on variables in $\exists(\beta|\alpha)[l, u]$. Its *correct answer under $s$* is Yes if such a $\theta$ exists and No otherwise. Given a probabilistic query $\exists(\beta|\alpha)[x, y]$ with $x, y \in \mathcal{X}$ to a probabilistic logic program $P$, its *tight answer substitutions under $s$* are substitutions $\theta$ such that $P \|\sim^s_{tight} (\beta\theta|\alpha\theta)[x\theta, y\theta]$, that $\theta$ acts only on variables in $\exists(\beta|\alpha)[x, y]$, and that $x\theta, y\theta \in [0, 1]$.

### 2.3 EXAMPLES

We now give some illustrative examples. In the first example, 1-entailment shows the property of inheritance of probabilistic knowledge, while 0-entailment does not.

**Example 2.1** The knowledge "all penguins are birds" and "birds have legs with a probability of at least 0.95" can be expressed by the following probabilistic logic program $P$:

$$P = \{(b(X)|p(X))[1, 1], (l(X)|b(X))[.95, 1]\}.$$

Our wondering about the tight interval for the probability that Tweety has legs given that Tweety is a penguin can be expressed by the object-ground probabilistic query

$$Q = \exists(l(tweety)|p(tweety))[R, S].$$

Its tight answer substitutions under 0- and 1-entailment are given by $\{R/0, S/1\}$ and $\{R/.95, S/1\}$, respectively, as

$P \models_{tight} (l(tweety)|p(tweety))[0, 1]$ and
$P \cup \{(p(tweety)|\top)[1, 1]\} \models_{tight} (l(tweety)|\top)[.95, 1]$.

The next example shows that inheritance in 1-entailment may often result in incompatible probabilistic knowledge, as 1-entailment does not have overriding mechanisms.

**Example 2.2** The knowledge "all penguins are birds", "birds have legs with a probability of at least 0.95", "birds fly with a probability between 0.9 and 0.95", and "penguins fly with a probability of at most 0.05" can be expressed by the following probabilistic logic program $P$:

$$P = \{(b(X)|p(X))[1, 1], (l(X)|b(X))[.95, 1],$$
$$(f(X)|b(X))[.9, .95], (f(X)|p(X))[0, .05]\}.$$

Our wondering about the tight interval for the probability that Tweety has legs given that Tweety is a penguin can be expressed by the object-ground probabilistic query

$$Q = \exists(l(tweety)|p(tweety))[R, S].$$

Its tight answer substitutions under 0- and 1-entailment are given by $\{R/0, S/1\}$ and $\{R/1, S/0\}$, respectively, as

$P \models_{tight} (l(tweety)|p(tweety))[0, 1]$ and
$P \cup \{(p(tweety)|\top)[1, 1]\} \models_{tight} (l(tweety)|\top)[1, 0]$.

Note that we obtain $[1, 0]$ under 1-entailment as $ground(P)$ does not have a model $Pr$ such that $Pr(p(tweety)) = 1$.

**Example 2.3** The knowledge "all magpies are birds", "birds chirp with a probability between 0.7 and 0.8", and "magpies chirp with a probability of at most 0.99" can be expressed by the following probabilistic logic program $P$:

$$P = \{(b(X)|m(X))[1, 1], (c(X)|b(X))[.7, .8],$$
$$(c(X)|m(X))[0, .99]\}.$$

Our wondering about the tight interval for the probability that Sam chirps given that Sam is a magpie can be expressed by the object-ground probabilistic query

$$Q = \exists(c(sam)|m(sam))[R, S].$$

Its tight answer substitutions under 0- and 1-entailment are given by $\{R/0, S/.99\}$ and $\{R/.7, S/.8\}$, respectively.

## 3 MOTIVATION AND KEY IDEAS

Under 0- and 1-entailment, classical conditional constraints $(\psi|\phi)[1,1]$ and $(\psi|\phi)[0,0]$ are interpreted as "$\phi$ implies $\psi$" and "$\phi$ implies $\neg\psi$", respectively. That is, both 0- and 1-entailment satisfy the following property of *inheritance of classical knowledge along subclass relationships*:

**IC** If $C \mathrel{\|\!\sim} (\psi|\phi)[c,c]$ and $\phi \Leftarrow \phi^*$ is logically valid, then $C \mathrel{\|\!\sim} (\psi|\phi^*)[c,c]$,

for all ground classical formulas $\psi$, $\phi$, and $\phi^*$, all sets of ground conditional constraints $C$, and all $c \in \{0,1\}$.

More generally, however, 0-entailment interprets conditional constraints $(\psi|\phi)[l,u]$ as "the conditional probability of $\psi$ given $\phi$ lies between $l$ and $u$". That is, 0-entailment does *not* have the following property of *inheritance of probabilistic knowledge along subclass relationships*:

**IP** If $C \mathrel{\|\!\sim} (\psi|\phi)[l,u]$ and $\phi \Leftarrow \phi^*$ is logically valid, then $C \mathrel{\|\!\sim} (\psi|\phi^*)[l,u]$.

for all ground classical formulas $\psi$, $\phi$, and $\phi^*$, all sets of ground conditional constraints $C$, and all $l, u \in [0,1]$.

Moreover, 1-entailment interprets $(\psi|\phi)[l,u]$ as "$\phi$ implies that $\psi$ holds with a probability between $l$ and $u$". That is, 1-entailment satisfies **IP**, but it does not realize overriding. As the inherited knowledge is often incompatible, we thus often conclude the empty interval (see Example 2.2).

In summary, 0-entailment does not have the property **IP**, while 1-entailment satisfies **IP**, but does not realize overriding. Inheritance with overriding, however, is a desirable feature of probabilistic entailment relations, which is well-known from *reference class reasoning* [34, 16, 17, 31].

A natural way to obtain inheritance with overriding is to weaken 1-entailment by interpreting purely probabilistic conditional constraints $(\psi|\phi)[l,u]$ as "*generally*, $\phi$ implies that $\psi$ holds with a probability between $l$ and $u$". We formalize this idea by using recent notions of entailment for probabilistic default theories [21, 22], which are based on default reasoning with conditional knowledge bases.

## 4 INHERITANCE WITH OVERRIDING

We briefly recall approaches to probabilistic default reasoning from [21, 22], which we then use to define entailment under inheritance with overriding. We consider only $z$- and $lex$-entailment here. Another approach to probabilistic default reasoning is $c$-entailment [21, 22], which can also be used for entailment under inheritance with overriding.

### 4.1 PRELIMINARIES

A *probabilistic default theory* $T = (S, D)$ is a pair of finite sets $S$ and $D$ of ground conditional constraints. The elements in $S$ and $D$ are called *strict conditional constraints* and *defeasible conditional constraints* (or also *defaults*), respectively. Intuitively, every strict conditional constraint $(\psi|\phi)[l,u] \in S$ is interpreted as "the conditional probability of $\psi$ given $\phi$ is between $l$ and $u$", while every default $(\psi|\phi)[l,u] \in D$ is interpreted as "generally, $\phi$ implies that $\psi$ holds with a probability between $l$ and $u$".

A probabilistic interpretation $Pr$ *verifies* a ground conditional constraints $(\psi|\phi)[l,u]$ iff $Pr(\phi) = 1$ and $Pr \models (\psi|\phi)[l,u]$. It *falsifies* $(\psi|\phi)[l,u]$ iff $Pr(\phi) = 1$ and $Pr \not\models (\psi|\phi)[l,u]$. A set of ground conditional constraints $D$ *tolerates* a ground conditional constraint $C$ *under* a set of ground conditional constraints $S$ iff $S \cup D$ has a model that verifies $C$. We say $D$ is *under $S$ in conflict* with $C$ iff no model of $S \cup D$ verifies $C$.

Given a probabilistic default theory $T = (S, D)$, a *default ranking* $\kappa$ on $D$ maps each $C \in D$ to a nonnegative integer. We say $\kappa$ is *admissible* with $T = (S, D)$ iff each $D' \subseteq D$ that is under $S$ in conflict with some $C \in D$ contains a conditional constraint $C'$ such that $\kappa(C') < \kappa(C)$. A probabilistic default theory $T = (S, D)$ is *consistent* iff there exists a default ranking on $D$ that is admissible with $T$. It is *inconsistent* iff no such default ranking exists.

A *probability ranking* $\kappa$ assigns to each probabilistic interpretation on $\mathcal{I}_\Phi$ a member of $\{0, 1, \ldots\} \cup \{\infty\}$ such that $\kappa(Pr) = 0$ for at least one interpretation $Pr$.

### 4.2 ENTAILMENT IN SYSTEM $Z$

We now recall $z$-entailment for probabilistic default theories [21, 22], which is a proper generalization of Pearl's entailment in system $Z$ [30, 11]. It is defined with respect to a consistent probabilistic default theory $T = (S, D)$.

The notion of $z$-entailment for probabilistic default theories is linked to an ordered partition of $D$, a default ranking $z$ on $D$, and a probability ranking $\kappa^z$.

Let $(D_0, \ldots, D_k)$ be the unique ordered partition of $D$ such that each $D_i$ is the set of all $d \in D - \bigcup\{D_j \mid 0 \leq j < i\}$ tolerated under $S$ by $D - \bigcup\{D_j \mid 0 \leq j < i\}$. We call this $(D_0, \ldots, D_k)$ the $z$-*partition* of $D$.

For every $j \in \{0, \ldots, k\}$, each $d \in D_j$ is assigned the value $j$ under the default ranking $z$. The probability ranking $\kappa^z$ on all probabilistic interpretations $Pr$ is then defined by

$$\kappa^z(Pr) = \begin{cases} \infty & \text{if } Pr \not\models S \\ 0 & \text{if } Pr \models S \cup D \\ 1 + \max_{d \in D: \ Pr \not\models d} z(d) & \text{otherwise.} \end{cases}$$

Note that $z$ is a default ranking admissible with $T$ [22].

The probability ranking $\kappa^z$ defines a preference relation on probabilistic interpretations as follows. For probabilistic interpretations $Pr$ and $Pr'$, we say $Pr$ is $z$-*preferable* to $Pr'$ iff $\kappa^z(Pr) < \kappa^z(Pr')$. A model $Pr$ of a set of ground probabilistic formulas $\mathcal{F}$ is a *z-minimal model* of $\mathcal{F}$ iff no model of $\mathcal{F}$ is *z*-preferable to $Pr$.

For ground probabilistic formulas $E$ and $F$, we say $F$ is a *z-consequence* of $E$, denoted $E \mid\!\sim^z F$, iff every *z*-minimal model of $S \cup \{E\}$ satisfies $F$. A ground conditional constraint $(\psi|\phi)[l,u]$ is a *tight z-consequence* of $E$, denoted $E \mid\!\sim^z_{tight} (\psi|\phi)[l,u]$, iff $l$ (resp., $u$) is the infimum (resp., supremum) of $Pr(\psi|\phi)$ subject to all *z*-minimal models $Pr$ of $S \cup \{E\}$ with $Pr(\phi) > 0$.

### 4.3 LEXICOGRAPHIC ENTAILMENT

We next recall *lex*-entailment for probabilistic default theories [21, 22], which is a proper generalization of Lehmann's lexicographic entailment [18]. In the sequel, consider a consistent probabilistic default theory $T = (S, D)$.

We use the *z*-partition $(D_0, \ldots, D_k)$ of $D$ to define a lexicographic preference relation on probabilistic interpretations as follows. For probabilistic interpretations $Pr$ and $Pr'$, we say $Pr$ is *lex-preferable* to $Pr'$ iff there is some $i \in \{0, \ldots, k\}$ such that $|\{d \in D_i \mid Pr \models d\}| > |\{d \in D_i \mid Pr' \models d\}|$ and $|\{d \in D_j \mid Pr \models d\}| = |\{d \in D_j \mid Pr' \models d\}|$ for all $i < j \leq k$. A model $Pr$ of a set of probabilistic formulas $\mathcal{F}$ is a *lex-minimal model* of $\mathcal{F}$ iff no model of $\mathcal{F}$ is *lex*-preferable to $Pr$.

For ground probabilistic formulas $E$ and $F$, we say $F$ is a *lex-consequence* of $E$, denoted $E \mid\!\sim^{lex} F$, iff every *lex*-minimal model of $S \cup \{E\}$ satisfies $F$. A ground conditional constraint $(\psi|\phi)[l,u]$ is a *tight lex-consequence* of $E$, denoted $E \mid\!\sim^{lex}_{tight} (\psi|\phi)[l,u]$, iff $l$ (resp., $u$) is the infimum (resp., supremum) of $Pr(\psi|\phi)$ subject to all *lex*-minimal models $Pr$ of $S \cup \{E\}$ with $Pr(\phi) > 0$.

### 4.4 INHERITANCE WITH OVERRIDING

We now introduce *z*- and *lex*-entailment for probabilistic logic programs, which are based on *z*- and *lex*-entailment for probabilistic default theories.

Each probabilistic logic program $P$ is associated with the probabilistic default theory $T(P) = (S(P), D(P))$, where $S(P)$ (resp., $D(P)$) is the set of all classical (resp., purely probabilistic) members of $ground(P)$. A probabilistic logic program $P$ is *consistent* iff $T(P)$ is consistent. The *z-partition* of $P$ is the *z*-partition of $T(P)$.

We now define the notion of *s-entailment*, $s \in \{z, lex\}$, for probabilistic logic programs $P$. A ground conditional constraint $C = (\psi|\phi)[l,u]$ is an *s-consequence* of $P$, denoted $P \mid\!\sim^s C$, iff $(\phi|\top)[1,1] \mid\!\sim^s (\psi|\top)[l,u]$ under $T(P)$. It is a *tight s-consequence* of $P$, denoted $P \mid\!\sim^s_{tight} C$, iff $(\phi|\top)[1,1] \mid\!\sim^s_{tight} (\psi|\top)[l,u]$ under $T(P)$. A conditional constraint $C = (\psi|\phi)[l,u]$ is an *s-consequence* of $P$, denoted $P \mid\!\sim^s C$, iff all ground instances of $C$ are *s*-consequences of $P$. It is a *tight s-consequence* of $P$, denoted $P \mid\!\sim^s_{tight} C$, iff $l$ (resp., $u$) is the infimum (resp., supremum) of $a$ (resp., $b$) subject to $P \mid\!\sim_{tight} (\psi'|\phi')[a,b]$ and all ground instances $(\psi'|\phi')$ of $(\psi|\phi)$.

### 4.5 EXAMPLES

We now give some illustrative examples.

**Example 4.1** Consider again the probabilistic logic program $P$ and the object-ground probabilistic query $Q$ in Example 2.1. The tight answer substitution for $Q$ to $P$ under both *z*- and *lex*-entailment is given by $\{R/.95, S/1\}$.

**Example 4.2** Consider the probabilistic logic program $P$ and the object-ground probabilistic query $Q$ in Example 2.2. The tight answer substitution for $Q$ to $P$ under both *z*- and *lex*-entailment is given by $\{R/.95, S/1\}$.

**Example 4.3** Consider again the probabilistic logic program $P$ and the object-ground probabilistic query $Q$ in Example 2.3. The tight answer substitution for $Q$ to $P$ under both *z*- and *lex*-entailment is given by $\{R/.7, S/.8\}$.

## 5 SEMANTIC PROPERTIES

We now analyze some general nonmonotonic properties of *z*- and *lex*-entailment for probabilistic logic programs.

In the sequel, we write $P \mid\!\sim (\phi|\varepsilon \vee \varepsilon')[l,u]$ to denote that $(\varepsilon|\top)[1,1] \vee (\varepsilon'|\top)[1,1] \mid\!\sim (\phi|\top)[l,u]$ under $T(P)$. We use $P \mid\!\not\sim C$ to denote that it is not the case that $P \mid\!\sim C$. We implicitly assume that all notions of entailment are naturally extended to negations of conditional constraints of the form $\neg(\beta|\alpha)[r,s]$, which are true in a probabilistic interpretation $Pr$ iff $Pr(\alpha) > 0$ and $Pr(\beta|\alpha) \notin [r,s]$.

We first consider the postulates *Right Weakening (RW)*, *Reflexivity (Ref)*, *Left Logical Equivalence (LLE)*, *Cut*, *Cautious Monotonicity (CM)*, and *Or* proposed by Kraus, Lehmann, and Magidor [15], which are commonly regarded as being particularly desirable for any reasonable notion of nonmonotonic entailment. The following result shows that *z*- and *lex*-entailment satisfy these postulates.

**Theorem 5.1** $\mid\!\sim^z$ and $\mid\!\sim^{lex}$ *satisfy the following properties for all probabilistic logic programs $P$, all ground classical formulas $\varepsilon, \varepsilon', \phi,$ and $\psi$, and all $l, l', u, u' \in [0,1]$:*

*RW.* If $(\phi|\top)[l,u] \Rightarrow (\psi|\top)[l',u']$ *is logically valid and* $P \mid\!\sim (\phi|\varepsilon)[l,u]$, *then* $P \mid\!\sim (\psi|\varepsilon)[l',u']$.

*Ref.* $P \mid\!\sim (\varepsilon|\varepsilon)[1,1]$.

*LLE.* If $\varepsilon \Leftrightarrow \varepsilon'$ *is logically valid, then* $P \mid\!\sim (\phi|\varepsilon)[l,u]$ *iff* $P \mid\!\sim (\phi|\varepsilon')[l,u]$.

*Cut.* If $P \mid\!\sim (\varepsilon|\varepsilon')[1,1]$ *and* $P \mid\!\sim (\phi|\varepsilon \wedge \varepsilon')[l,u]$, *then* $P \mid\!\sim (\phi|\varepsilon')[l,u]$.

*CM.* If $P \mid\!\sim (\varepsilon|\varepsilon')[1,1]$ *and* $P \mid\!\sim (\phi|\varepsilon')[l,u]$, *then* $P \mid\!\sim (\phi|\varepsilon \wedge \varepsilon')[l,u]$.

*Or.* If $P \mathrel{|\!\sim} (\phi|\varepsilon)[l,u]$ and $P \mathrel{|\!\sim} (\phi|\varepsilon')[l,u]$,
then $P \mathrel{|\!\sim} (\phi|\varepsilon \underline{\vee} \varepsilon')[l,u]$.

Another desirable property is *Rational Monotonicity (RM)* [15], which describes a restricted form of monotony, and allows to ignore certain kinds of irrelevant knowledge. The next result shows that $z$- and $lex$-entailment satisfy *RM*.

**Theorem 5.2** $\mathrel{|\!\sim}^z$ and $\mathrel{|\!\sim}^{lex}$ *satisfy the following property for all probabilistic logic programs $P$, all ground classical formulas $\varepsilon$, $\varepsilon'$, and $\psi$, and all $l, u \in [0,1]$:*

*RM.* If $P \mathrel{|\!\sim} (\psi|\varepsilon)[l,u]$ and $P \mathrel{|\!\not\sim} \neg(\varepsilon'|\varepsilon)[1,1]$, then $P \mathrel{|\!\sim} (\psi|\varepsilon \wedge \varepsilon')[l,u]$.

We finally consider the properties *Irrelevance (Irr)* and *Direct Inference (DI)*, which are adapted from [2] and [1], respectively. Informally, *Irr* says that $\varepsilon'$ is irrelevant to a conclusion "$P \mathrel{|\!\sim} (\psi|\varepsilon)[l,u]$" when they are defined over disjoint sets of atoms. While *DI* expresses that $P$ should entail all its own conditional constraints. The following result shows that $z$- and $lex$-entailment satisfy *Irr* and *DI*.

**Theorem 5.3** $\mathrel{|\!\sim}^z$ and $\mathrel{|\!\sim}^{lex}$ *satisfy the following properties for all probabilistic logic programs $P$, all ground classical formulas $\varepsilon$, $\varepsilon'$, $\phi$, and $\psi$, and all $l, u \in [0,1]$:*

*Irr.* If $P \mathrel{|\!\sim} (\psi|\varepsilon)[l,u]$, and no atom of $ground(P)$ and $(\psi|\varepsilon)[l,u]$ occurs in $\varepsilon'$, then $P \mathrel{|\!\sim} (\psi|\varepsilon \wedge \varepsilon')[l,u]$.

*DI.* If $(\psi|\phi)[l,u] \in ground(P)$ and $\varepsilon \Leftrightarrow \phi$ is logically valid, then $P \mathrel{|\!\sim} (\psi|\varepsilon)[l,u]$.

Note that entailment under inheritance with overriding based on $c$-entailment [21, 22] satisfies all the above properties except for Rational Monotonicity.

## 6 ALGORITHMS

In this section, we give algorithms for probabilistic logic programming under inheritance with overriding.

### 6.1 OVERVIEW

We consider the following problems:

CONSISTENCY: Given a probabilistic logic program $P$, decide whether $P$ is consistent.

TIGHT S-CONSEQUENCE: Given a consistent probabilistic logic program $P$ and an object-ground probabilistic query $Q = \exists(\beta|\alpha)[x,y]$, compute the tight answer substitution for $Q$ to $P$ under some fixed $s \in \{z, lex\}$.

The main idea behind our algorithms is to reduce these problems to the problem of deciding whether a probabilistic logic program is satisfiable and the problem of computing the tight answer substitution for an object-ground probabilistic query to a probabilistic logic program under 0-entailment, which we denote SATISFIABILITY and TIGHT LOGICAL CONSEQUENCE, respectively.

### 6.2 CONSISTENCY

Algorithm $z$-partition (see Fig. 1) decides whether a probabilistic logic program $P$ is consistent. If this is the case, then $z$-partition returns the $z$-partition of $P$, otherwise *nil*. Note that in Step 5 of $z$-partition, a number of instances of SATISFIABILITY must be solved. Algorithm $z$-partition is essentially a reformulation of an algorithm for deciding $\varepsilon$-consistency in default reasoning from conditional knowledge bases by Goldszmidt and Pearl [10].

---

**Algorithm $z$-partition (essentially [10])**

**Input**: Probabilistic logic program $P$ with $D(P) \neq \emptyset$.
**Output**: $z$-partition of $P$, if $P$ is consistent, otherwise *nil*.

1. $R := D(P)$;
2. $i := -1$;
3. **repeat**
4.    $i := i + 1$
5.    $D[i] := \{C \in R \mid C \text{ is tolerated under } S(P) \text{ by } R\}$;
6.    $R := R - D[i]$
7. **until** $R = \emptyset$ or $D[i] = \emptyset$;
8. **if** $R = \emptyset$ **then return** $(D[0], \ldots, D[i])$
9.    **else return** *nil*.

---

Figure 1: Algorithm $z$-partition

### 6.3 TIGHT S-CONSEQUENCE

In the sequel, let $P$ be a consistent probabilistic logic program, and let $(D_0, \ldots, D_k)$ be its $z$-partition.

For $G, H \subseteq D(P)$, we say $G$ is *$z$-preferable* to $H$ iff some $i \in \{0, \ldots, k\}$ exists such that $D_i \subseteq G$, $D_i \not\subseteq H$, and $D_j \subseteq G$ and $D_j \subseteq H$ for all $i < j \leq k$. We say $G$ is *lex-preferable* to $H$ iff some $i \in \{0, \ldots, k\}$ exists such that $|G \cap D_i| > |H \cap D_i|$ and $|G \cap D_j| = |H \cap D_j|$ for all $i < j \leq k$. For $\mathcal{D} \subseteq 2^{D(P)}$ and $s \in \{z, lex\}$, we say $G$ is *$s$-minimal* in $\mathcal{D}$ iff $G \in \mathcal{D}$ and no $H \in \mathcal{D}$ is $s$-preferable to $G$.

We now reduce TIGHT S-CONSEQUENCE to SATISFIABILITY and TIGHT LOGICAL CONSEQUENCE. The key idea behind this reduction is that there exists a set $\mathcal{D}_\alpha^s(P)$ of subsets of $D(P)$ such that $P \mathrel{|\!\sim}^s (\beta|\alpha)[l,u]$ iff $S(P) \cup H \cup \{(\alpha|\top)[1,1]\} \models (\beta|\top)[l,u]$ for all $H \in \mathcal{D}_\alpha^s(P)$.

**Theorem 6.1** *Let $P$ be a consistent probabilistic logic program, and let $\exists(\beta|\alpha)[x,y]$ be an object-ground probabilistic query. Let $s \in \{z, lex\}$. Let $R = S(P) \cup \{(\alpha|\top)[1,1]\}$, and let $\mathcal{D}_\alpha^s(P)$ be the set of all $s$-minimal elements in $\{H \subseteq D(P) \mid R \cup H \text{ is satisfiable}\}$. Then, $l$ (resp., $u$) such that $P \mathrel{|\!\sim}^s_{tight} (\beta|\alpha)[l,u]$ is given as follows:*

*(a) If $R$ is unsatisfiable, then $l = 1$ (resp., $u = 0$).*

*(b) Otherwise, $l = \min c$ (resp., $u = \max d$) subject to $R \cup H \models_{tight} (\beta|\top)[c,d]$ and $H \in \mathcal{D}_\alpha^s(P)$.*

Based on Theorem 6.1, Algorithm tight-$z$-consequence (resp., tight-$lex$-consequence) computes tight answer substitutions under $z$-entailment (resp., $lex$-entailment).

Step 2 checks whether $R$ is unsatisfiable. If this is the case, then $\theta = \{x/1, y/0\}$ is returned by Theorem 6.1 (a). Otherwise, we compute $\mathcal{D}_\alpha^s(P)$ along the $z$-partition of $P$ in steps 3–7 (resp., steps 3–15), and the tight answer substitution using Theorem 6.1 (b) in step 8 (resp., steps 16–20).

---

**Algorithm tight-$z$-consequence**

**Input**: Consistent probabilistic logic program $P$ and object-ground probabilistic query $Q = \exists(\beta|\alpha)[x, y]$.
**Output**: Tight answer substitution $\theta = \{x/l, y/u\}$ for $Q$ to $P$ under $z$-entailment.
Notation: $(D_0, \ldots, D_k)$ denotes the $z$-partition of $T(P)$.

1. $R := S(P) \cup \{(\alpha|\top)[1,1]\}$;
2. **if** $R$ is unsatisfiable **then return** $\theta = \{x/1, y/0\}$;
3. $j := k$;
4. **while** $j \geq 0$ and $R \cup D_j$ is satisfiable **do begin**
5. 　$R := R \cup D_j$;
6. 　$j := j - 1$;
7. **end**;
8. compute $l, u \in [0, 1]$ such that $R \models_{tight} (\beta|\top)[l, u]$;
9. **return** $\theta = \{x/l, y/u\}$.

---

Figure 2: Algorithm tight-$z$-consequence

---

**Algorithm tight-$lex$-consequence**

**Input**: Consistent probabilistic logic program $P$ and object-ground probabilistic query $Q = \exists(\beta|\alpha)[x, y]$.
**Output**: Tight answer substitution $\theta = \{x/l, y/u\}$ for $Q$ to $P$ under $lex$-entailment.
Notation: $(D_0, \ldots, D_k)$ denotes the $z$-partition of $T(P)$.

1. $R := S(P) \cup \{(\alpha|\top)[1,1]\}$;
2. **if** $R$ is unsatisfiable **then return** $\theta = \{x/1, y/0\}$;
3. $\mathcal{H} := \{\emptyset\}$;
4. **for** $j := k$ **downto** 0 **do begin**
5. 　$n := 0$;
6. 　$\mathcal{H}' := \emptyset$;
7. 　**for each** $G \subseteq D_j$ and $H \in \mathcal{H}$ **do**
8. 　　**if** $R \cup G \cup H$ is satisfiable **then**
9. 　　　**if** $n = |G|$ **then** $\mathcal{H}' := \mathcal{H}' \cup \{G \cup H\}$
10. 　　　**else if** $n < |G|$ **then begin**
11. 　　　　$\mathcal{H}' := \{G \cup H\}$;
12. 　　　　$n := |G|$
13. 　　　**end**;
14. 　$\mathcal{H} := \mathcal{H}'$;
15. **end**;
16. $(l, u) := (1, 0)$;
17. **for each** $H \in \mathcal{H}$ **do begin**
18. 　compute $c, d \in [0, 1]$ s.t. $R \cup H \models_{tight} (\beta|\top)[c, d]$;
19. 　$(l, u) := (\min(l, c), \max(u, d))$
20. **end**;
21. **return** $\theta = \{x/l, y/u\}$.

---

Figure 3: Algorithm tight-$lex$-consequence

## 7　COMPUTATIONAL COMPLEXITY

In this section, we characterize the computational complexity of the problems CONSISTENCY and TIGHT S-CONSEQUENCE in the propositional case.

### 7.1　COMPLEXITY CLASSES

We briefly describe the complexity classes that occur in our complexity results. See [9, 29] for further background.

The class NP contains all decision problems that can be solved in nondeterministic polynomial time. The class $\Delta_2^P$ contains all decision problems that can be solved in deterministic polynomial time with an oracle for NP.

To classify problems that compute an output value, rather than a Yes/No-answer, function classes have been introduced. In particular, $F\Delta_2^P$ is the functional analog to $\Delta_2^P$.

### 7.2　COMPLEXITY RESULTS

We consider the general propositional case as well as the restriction to the 1-conjunctive propositional case. In both cases, the given probabilistic logic programs $P$ are ground and the given probabilistic queries $Q$ are object-ground. In the 1-conjunctive propositional case, we then additionally assume that $P$ and $Q$ are 1-conjunctive.

The results on the propositional complexity of CONSISTENCY and TIGHT S-CONSEQUENCE are shown in Tables 1–2. In detail, CONSISTENCY and TIGHT S-CONSEQUENCE are NP- and $F\Delta_2^P$-complete, respectively, in the general and the 1-conjunctive propositional case.

That is, they have the same complexity as the problems SATISFIABILITY and TIGHT LOGICAL CONSEQUENCE, respectively, in the respective propositional cases [19]. Intuitively, adding inheritance with overriding to probabilistic logic programming does not increase its complexity.

Table 1: Prop. Complexity of CONSISTENCY

|  | general case | 1-conjunctive case |
| --- | --- | --- |
| consistency | NP-complete | NP-complete |

Table 2: Prop. Complexity of TIGHT S-CONSEQUENCE

|  | general case | 1-conjunctive case |
| --- | --- | --- |
| $\models_{tight}^z$ | $F\Delta_2^P$-complete | $F\Delta_2^P$-complete |
| $\models_{tight}^{lex}$ | $F\Delta_2^P$-complete | $F\Delta_2^P$-complete |

## 8　SUMMARY AND OUTLOOK

We presented probabilistic logic programming under inheritance with overriding, which is based on recent approaches to probabilistic default reasoning. We described some general nonmonotonic properties of entailment under inheritance with overriding. Moreover, we presented algorithms for probabilistic logic programming under inheritance with overriding, and we analyzed its propositional complexity.

A very interesting topic of future research is to investigate the relationship to probabilistic logic programming under maximum entropy as presented in [24], where we also have some form of inheritance with overriding.

**Acknowledgements**

I am very grateful to Gabriele Kern-Isberner for valuable comments on an earlier version of this paper. Many thanks also the anonymous reviewers for their useful comments. This work has been supported by a DFG grant and the Austrian Science Fund under project N Z29-INF.